\documentclass[10pt,twocolumn,letterpaper]{article}
\usepackage{cvpr}
\usepackage[T1]{fontenc}
\usepackage{amsmath,amssymb}
\usepackage{graphicx,verbatim}
\usepackage{booktabs}
\usepackage{multirow}
\usepackage{makecell}
\usepackage[table]{xcolor}
\usepackage{colortbl}
\usepackage{caption}
\usepackage{float}
\usepackage{subcaption}
\usepackage{placeins}
\usepackage{xcolor}

\definecolor{cvprblue}{rgb}{0.21,0.49,0.74}
\usepackage[pagebackref,breaklinks,colorlinks,allcolors=cvprblue]{hyperref}
\definecolor{OursPurple}{RGB}{128, 90, 213}

\def\paperID{29}
\def\confName{CVPR}
\def\confYear{2026}

\title{BiomedAP: A Vision-Informed Dual-Anchor Framework with Gated Cross-Modal Fusion for Robust Medical Vision-Language Adaptation}
\author{
Huanyang Tong\textsuperscript{*}\\
Wenzhou University\\
{\small \texttt{huanyangtong@stu.wzu.edu.cn}}
\and
Kai Liu\textsuperscript{\dag}\\
Wenzhou University\\
{\small \texttt{kailiu@wzu.edu.cn}}
\and
Fangjun Kuang\textsuperscript{\dag}\\
Wenzhou Business College\\
{\small \texttt{kfj@wzbc.edu.cn}}
\and
Huiling Chen\textsuperscript{\dag}\\
Wenzhou University\\
{\small \texttt{chenhuiling\_jsj@wzu.edu.cn}}
}

\begin{document}
\maketitle

\begingroup
\renewcommand\thefootnote{}
\footnotetext{
* First author \quad
\dag Corresponding author
}
\endgroup

\begin{abstract}

  Biomedical Vision--Language Models (VLMs) have shown remarkable promise in few-shot medical diagnosis but face a critical bottleneck: \textit{fragility to prompt variations}.Existing adaptation frameworks typically optimize visual and textual prompts as independent streams, relying on ideal ``Golden Prompts''. In clinical reality, where descriptions are often noisy and heterogeneous, this modality isolation leads to unstable cross-modal alignment.

  To address this, we propose BiomedAP, a vision-informed dual-anchor framework with gated cross-modal fusion.BiomedAP enforces synergistic alignment through two mechanisms: (1) Gated Cross-Modal Fusion, which enables layer-wise interaction between modalities, acting as a dynamic noise regulator to suppress irrelevant textual cues; and (2) a Dual-Anchor Constraint that regularizes learnable prompts toward stable semantic centroids derived from both expert templates (High Anchors) and few-shot visual prototypes (Low Anchors).

  Extensive experiments across 11 benchmarks demonstrate that BiomedAP consistently surpasses baselines, achieving competitive few-shot accuracy and markedly enhanced robustness under prompt perturbations.

  Our code is available at: \url{https://github.com/tongdiedie/BiomedAP}.


\end{abstract}

\begin{figure*}[t]
  \centering
  \includegraphics[width=1.0\linewidth]{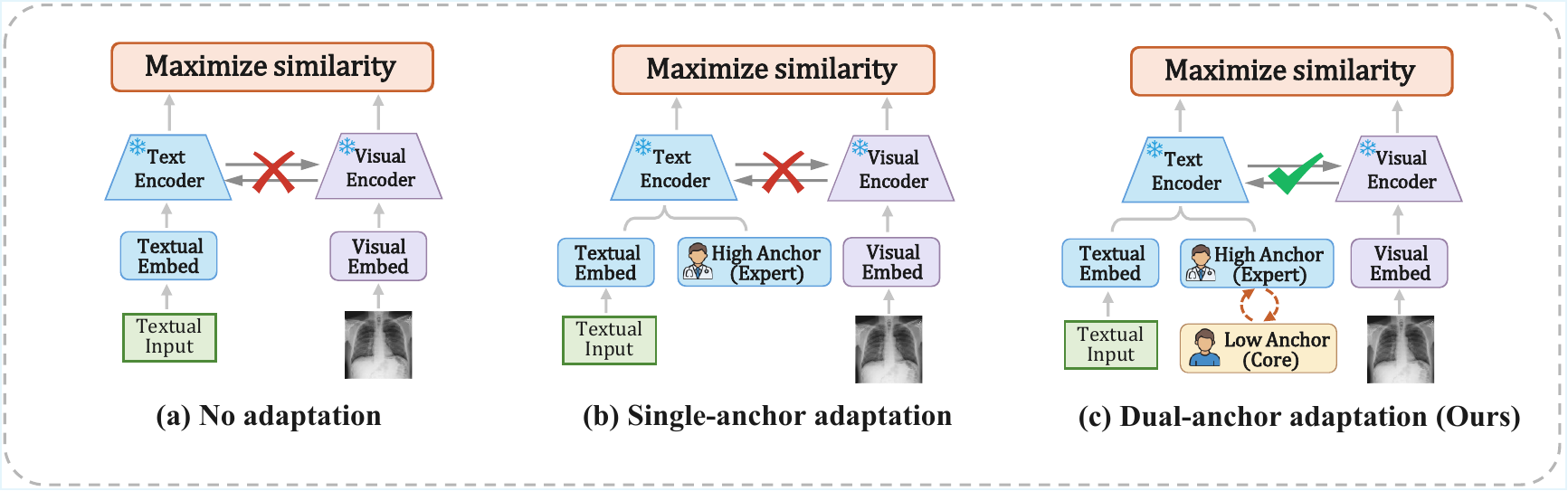}
  \caption{\textbf{Comparison of adaptation strategies.} 
  (a) \textbf{No Adaptation}: Direct inference without task-specific alignment.
  (b) \textbf{Single-Anchor Adaptation}: Relies solely on elaborate expert descriptions (High Anchor), leading to overfitting and fragility under prompt variations.
  (c) \textbf{Dual-Anchor Adaptation (Ours)}: Integrates both expert knowledge (High Anchor) and \textbf{vision-intrinsic semantics} derived from visual prototypes (Low Anchor), ensuring semantic stability across diverse clinical inputs.}
  \label{fig:motivation}
\end{figure*}

\section{Introduction}
Vision--language integration has emerged as a paradigm shift in medical AI \cite{moor2023medflamingo,wang2022medclip,zhang2023pmcvlm}.
Foundation VLMs, such as CLIP \cite{radford2021learning} and BiomedCLIP \cite{zhang2023biomedclip}, align images with text to enable flexible recognition across heterogeneous diseases.
Two mainstream adaptation lines are commonly used: context-token prompt learning (CoOp/CoCoOp) \cite{zhou2022conditional,zhou2022coop} and prompt-/adapter-style PEFT (e.g., VPT, Tip-Adapter, MaPLe) \cite{jia2022visual,khattak2023maple,zhang2022tip}, with biomedical variants such as Biomed-DPT \cite{peng2025biomeddpt}.
These methods are attractive because they retain most pretrained knowledge while requiring only a small number of trainable parameters, making them especially suitable for data-limited medical settings.

However, adapting these models to specific clinical tasks remains challenging.
Full fine-tuning is computationally prohibitive and prone to overfitting \cite{houlsby2019parameter,hu2021lora}, and standard PEFT often falls short of the reliability required for medical diagnosis \cite{liu2023adapterbias,zhou2024foundationmed}.
In real-world clinical settings, this challenge is further compounded by two recurring limitations.
\textbf{(1) Modality isolation limits semantic grounding.} Many pipelines rely on late-stage matching, coupling visual and textual representations only at the final similarity computation \cite{peng2025biomeddpt}.
This lack of intermediate interaction can amplify background confounders or miss subtle lesion patterns (Figure~\ref{fig:motivation}(a--b)).
\textbf{(2) Prompt fragility from single-anchor supervision.} Existing methods typically treat expert-crafted ``Golden Prompts'' as the sole optimization target.
Optimizing exclusively against a \textit{High Anchor} entangles semantics with surface forms, leading to brittle performance when clinical narratives become concise, noisy, or deviate from expert phrasing.

To overcome these challenges, we propose \textbf{BiomedAP} to learn \textit{invariant diagnostic semantics} from heterogeneous text.
As illustrated in Figure~\ref{fig:motivation}(c), BiomedAP introduces \textbf{Gated Cross-Modal Fusion} to deepen vision--text interaction during encoding.
A lightweight gate modulates textual injection based on consistency with visual evidence.
We further impose a \textbf{Dual-Anchor Constraint} with two centroids:
a \textbf{High Anchor (Expert)} for diagnostic specificity and a \textbf{Low Anchor (Core)} capturing \textbf{vision-informed intrinsic semantics}.
The Low Anchor stabilizes noisy prompts, while the High Anchor preserves fine-grained clinical details.

In summary, our main contributions are:
\begin{itemize}
    \item We identify \textit{modality isolation} and \textit{single-anchor fragility} as primary obstacles to robust biomedical PEFT.
    \item We introduce \textbf{BiomedAP} with \textbf{Gated Cross-Modal Fusion} to enable layer-wise interaction and suppress prompt-induced noise.
    \item We propose a \textbf{Dual-Anchor Constraint} that regularizes prompts toward both expert and \textbf{vision-informed} semantics, improving stability.
    \item Experiments on 11 benchmarks demonstrate consistent gains in few-shot learning and base-to-novel generalization, with superior robustness to prompt perturbations.
\end{itemize}

\section{Related Work} 

\noindent\textbf{\textit{Parameter-Efficient Adaptation of Vision-Language Models.}}
Parameter-efficient fine-tuning (PEFT) adapts large vision-language models (VLMs) with only a small number of trainable parameters \cite{gao2024clip,hu2021lora,radford2021learning,zhang2023biomedclip}.
Textual prompt learning methods, such as CoOp and CoCoOp, optimize learnable context tokens in the language branch \cite{zhou2022coop,zhou2022conditional}, whereas visual prompt tuning methods inject trainable tokens into the image encoder \cite{jia2022visual,yao2023kgcoop,zhu2023prograd}.
Recent work has further explored richer multimodal prompt parameterizations and improved generalization objectives for efficient VLM adaptation \cite{khattak2023maple,jie2024memvp,kim2025bayesian,liu2025surpl}.
However, many PEFT methods still adapt visual and textual streams in a largely decoupled manner, with cross-modal interaction occurring only at the final matching stage \cite{radford2021learning}.
Recent work has begun to explore more tightly coupled visual-textual prompt optimization, suggesting that stronger mutual interaction can improve cross-modal alignment \cite{mutualprompt2024vlm}.
This limitation is particularly restrictive in biomedical recognition, where reliable prediction often depends on tight alignment between localized visual evidence and clinical semantics \cite{boecking2022making,huang2021gloria}.

\noindent\textbf{\textit{Biomedical Prompt Learning and Medical VLM Adaptation.}}
Recent biomedical adaptation methods extend prompt learning to medical tasks using domain-specific corpora, expert templates, or LLM-generated descriptions, as broadly summarized in recent surveys of medical vision-language modeling \cite{lin2025taming}.
Recent work has also begun to explicitly study few-shot adaptation as a dedicated evaluation setting for medical vision-language models \cite{shakeri2024fewshotmedvlm}.
Methods such as BiomedCoOp and Biomed-DPT show that biomedical prompt tuning substantially improves few-shot transfer over generic prompting baselines \cite{Koleilat_2025_CVPR,peng2025biomeddpt}.
Related biomedical VLMs, including MedCLIP, Med-Flamingo, and BiomedCLIP, further demonstrate the value of domain-aligned pre-training for medical vision-language reasoning \cite{wang2022medclip,moor2023medflamingo,zhang2023biomedclip}.
Nevertheless, most existing biomedical prompt learners still rely on late interaction and optimize prompts primarily in the text space, making them vulnerable to prompt fragility when clinical descriptions are noisy, concise, or phrased differently from training templates.
This concern is consistent with recent studies on robustness in medical vision-language models \cite{hussein2024promptsmooth}.

\noindent\textbf{\textit{Cross-Modal Interaction for Robust Grounding.}}
A complementary line of work improves vision-language alignment through deeper cross-modal interaction.
Representative models such as BLIP and BLIP-2 employ cross-attention and query-based fusion to strengthen multimodal grounding beyond late-stage matching \cite{li2022blip,li2023blip2}, and recent parameter-efficient multimodal learning methods similarly emphasize adaptive feature exchange across modalities \cite{Liang_2024_CVPR}.
In contrast to these approaches, BiomedAP is designed for few-shot biomedical adaptation under heterogeneous clinical text.
It combines intermediate-layer gated cross-modal fusion with dual-anchor regularization, enabling prompt learning to be both visually grounded and semantically stabilized.

\section{Method} 

\subsection{Overview}
\begin{figure*}[t]
  \centering
  \includegraphics[width=1.0\linewidth]{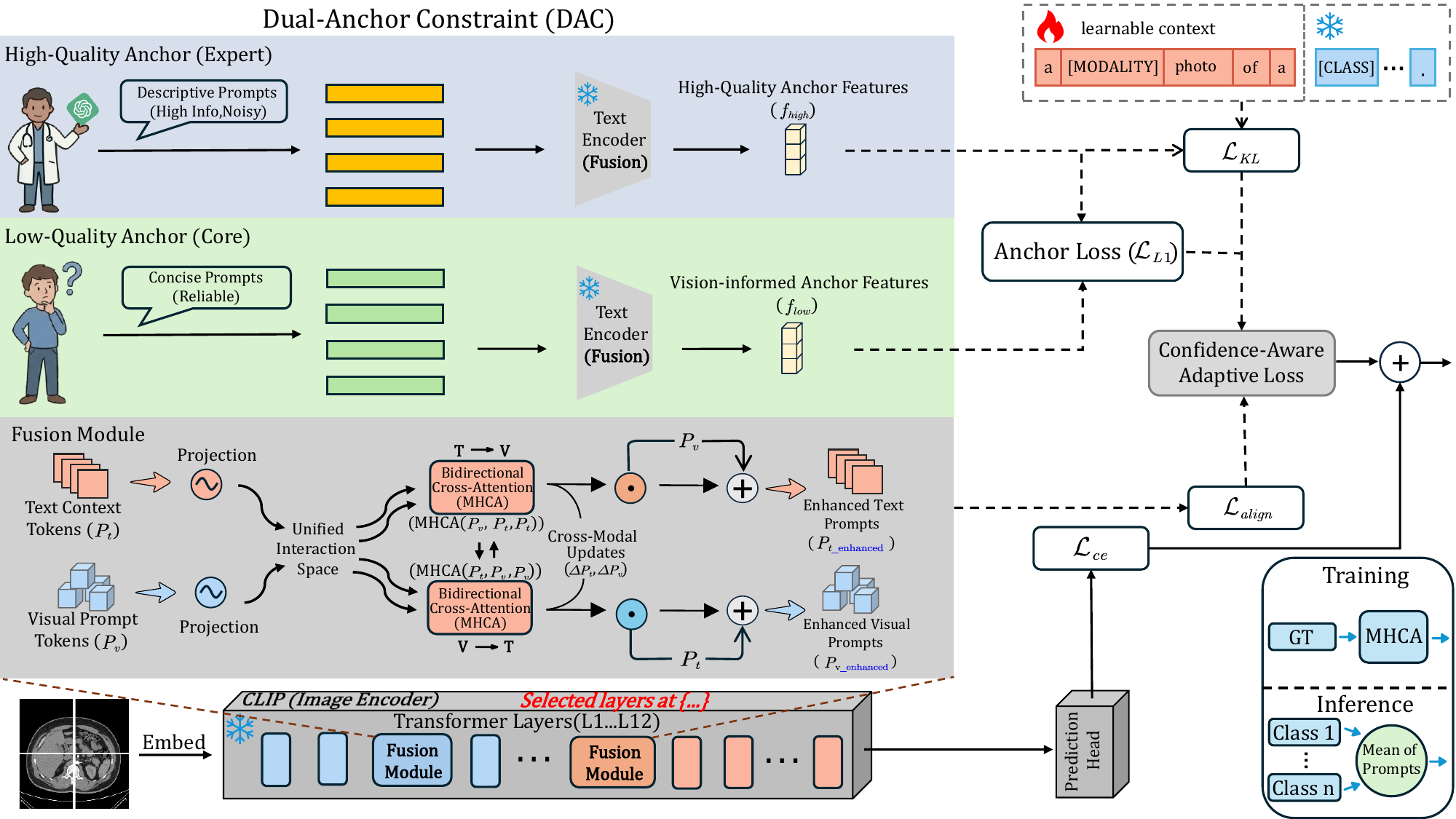}
  \caption{\textbf{Overall framework of BiomedAP.}
  DAC constructs a High-Quality (Expert) anchor and a Low-Quality (Core) anchor, whose features ($f_{high}$, $f_{low}$) regularize the learnable text context to reduce prompt drift.
  Training uses a \textit{Confidence-Aware Adaptive Loss} to balance anchor regularization with the task loss.
  A \textit{Unified Interaction Space} applies bidirectional MHCA between text-context and visual-prompt tokens at selected layers, followed by class-wise aggregation at inference.}
  \label{fig:architecture}
\end{figure*}

We introduce \textbf{BiomedAP}, a parameter-efficient framework that adapts frozen vision--language backbones to medical tasks \cite{houlsby2019parameter,hu2021lora}. 
BiomedAP addresses two common failure modes: (i) \textit{modality isolation} and (ii) \textit{prompt fragility}. 
As illustrated in Figure~\ref{fig:architecture}, we keep the pre-trained encoders frozen and optimize two lightweight components:
(1) \textbf{Gated Cross-Modal Prompt Fusion} for context-aware prompt refinement, and 
(2) \textbf{Dual-Anchor Constraint (DAC)} for semantic regularization.

\subsection{Gated Cross-Modal Prompt Fusion}
Let $\mathbf{P}_{v}^{l}\in\mathbb{R}^{M_{v}\times D_{v}}$ denote the visual prompt tokens injected into the frozen image encoder at layer $l$ \cite{jia2022visual}, and let $\mathbf{P}_{t}\in\mathbb{R}^{M_{t}\times D_{t}}$ denote learnable textual context tokens from the prompt learner \cite{zhou2022coop}.
Fusion modules are inserted at a small set of layers $L_f$ (e.g., \{5,8\}) to enable text-conditioned refinement of visual prompts while limiting negative transfer.

\paragraph{Projection to a unified interaction space.}
We project both prompts into a shared interaction space with $D_{\mathrm{uni}}=\max(D_{v},D_{t})$:
\begin{equation}
\tilde{\mathbf{P}}_{v}^{l}=\mathbf{P}_{v}^{l}\mathbf{W}_{v}^{in},\quad
\tilde{\mathbf{P}}_{t}^{l}=\mathbf{P}_{t}^{l}\mathbf{W}_{t}^{in},
\end{equation}
with learnable $\mathbf{W}_{v}^{in}$ and $\mathbf{W}_{t}^{in}$.

\paragraph{Bidirectional Cross-Attention.}
We use multi-head cross-attention (MHCA) to compute cross-modal updates:
\begin{equation}
\varDelta\mathbf{P}_{v}^{l}=\operatorname{MHCA}(\tilde{\mathbf{P}}_{v}^{l},\tilde{\mathbf{P}}_{t}^{l},\tilde{\mathbf{P}}_{t}^{l})\mathbf{W}_{v}^{out},
\end{equation}
where textual features serve as Key/Value to refine visual Queries; a symmetric operation updates $\mathbf{P}_{t}^{l}$.

\noindent\textbf{Gated Residual Refinement.}
To filter potentially harmful updates from noisy prompts, we condition the gate on both the current visual state $\mathbf{P}_{v}^{l}$ and the proposed update $\Delta\mathbf{P}_{v}^{l}$:
\begin{equation}
\alpha_{v}^{l} = \sigma([\mathbf{P}_{v}^{l} \,;\, \Delta\mathbf{P}_{v}^{l}] \mathbf{W}_{g}), \quad
\hat{\mathbf{P}}_{v}^{l} = \mathbf{P}_{v}^{l} + \alpha_{v}^{l} \odot \Delta\mathbf{P}_{v}^{l},
\end{equation}
where $[\cdot\,;\,\cdot]$ denotes concatenation, $\mathbf{W}_{g}$ is learnable, and $\sigma$ is Sigmoid.

\noindent\textbf{Intuition: Consistency-based Filtering.}
The gate suppresses incongruent cross-modal updates (e.g., induced by prompt noise) by driving $\alpha_{v}^{l}\!\to\!0$, while preserving and amplifying consistent updates that enhance diagnostic cues.

\paragraph{Inference with Global Medical Context.}
During training, fusion is conditioned on the ground-truth class prompt.
At inference, we compute a \textbf{Global Medical Context} by averaging the prompt embeddings of all candidate classes \cite{radford2021learning}, enabling label-free fusion without biasing toward a specific class.

\subsection{Dual-Anchor Constraint (DAC)}
Clinical descriptions exhibit substantial variability. To enforce semantic stability, we construct two complementary anchors that bridge the modality gap from opposite directions:

\begin{itemize}
    \item \textbf{High-quality Anchor} $\mathbf{f}_{high}$: An ensemble of expert-refined textual templates encoded by the frozen text encoder, capturing fine-grained medical knowledge (Top-down).
    \item \textbf{Low-quality Anchor} $\mathbf{f}_{low}^{c}$: A class-wise \emph{visual prototype} capturing intrinsic data semantics (Bottom-up). It is computed by averaging the features of few-shot support images $\{x_i^c\}_{i=1}^K$ using the frozen image encoder:
    \begin{equation}
    \mathbf{f}_{low}^{c} = \frac{1}{K}\sum_{i=1}^{K} \frac{\mathrm{Enc}_{img}(x_i^{c})}{\|\mathrm{Enc}_{img}(x_i^{c})\|_2}.
    \end{equation}
\end{itemize}

\noindent We regularize the learnable prompt feature $\mathbf{f}_{text}$ (associated with the ground-truth class $c$) to align with both anchors:
\begin{equation}
\mathcal{L}_{anchor}=\lambda_{high}\|\mathbf{f}_{text}-\mathbf{f}_{high}\|_{1}+\lambda_{low}\|\mathbf{f}_{text}-\mathbf{f}_{low}^{c}\|_{1}.
\label{eq:anchor}
\end{equation}

\noindent\textbf{Dual-Modality Centroids.} 
Standard prompt learning stays in the text space and can drift from visual distributions. By anchoring $\mathbf{f}_{text}$ to the \textbf{Visual Low Anchor}, DAC pulls prompts into a shared vision–language subspace, making the learned semantics both linguistically valid and visually grounded.

\subsection{Optimization Objective}
To reduce overfitting under \textbf{heterogeneous clinical supervision}, we optimize a composite objective:
\begin{equation}
\mathcal{L}_{\mathrm{total}}=\mathcal{L}_{\mathrm{CE}}(\mathbf{s},\mathbf{y}) + \mathcal{L}_{\mathrm{conf}}.
\end{equation}

\noindent\textbf{Confidence-aware Weighting.}
Standard static regularization ignores the reliability of supervision. To address this, we introduce an adaptive mechanism that strengthens regularization when the model's prediction is uncertain.
Let $\mathbf{p}=\mathrm{softmax}(\mathbf{s})$ denote the predicted probability distribution over $C$ classes. We quantify prediction confidence using normalized entropy:
\begin{equation}
\mathrm{conf} = 1 - \frac{H(\mathbf{p})}{\log C} \in [0,1], \quad \text{where } H(\mathbf{p}) = -\sum_{c=1}^{C} p_c \log p_c.
\end{equation}
We define an adaptive weight $w = 1 - \mathrm{conf}$ to dynamically scale the regularization intensity. The \textbf{Confidence-aware Adaptive Loss} $(\mathcal{L}_{\mathrm{conf}})$ is formulated as:
\begin{equation}
\mathcal{L}_{\mathrm{conf}}
= \lambda_{\mathrm{KD}}\,w\,\mathcal{L}_{\mathrm{KL}}
+ w\,\mathcal{L}_{\mathrm{anchor}}
+ \lambda_{\mathrm{align}}\,w\,\mathcal{L}_{\mathrm{align}}.
\end{equation}
Here, $\mathcal{L}_{\mathrm{KL}}$ preserves knowledge from the frozen backbone by distilling the adapted logits to match the frozen model outputs (with temperature $T$). 
Specifically, we compute $\mathcal{L}_{\mathrm{KL}}=\mathrm{KL}\!\left(\mathrm{softmax}(\mathbf{s}/T)\,\|\,\mathrm{softmax}(\mathbf{s}^{0}/T)\right)$, where $\mathbf{s}^{0}$ denotes the logits of the frozen backbone.
In addition, $\mathcal{L}_{\mathrm{align}}$ encourages cross-modal consistency by aligning the fused visual representation with the corresponding text prompt representation using a cosine similarity objective.

\section{Experiments}

\subsection{Experimental Setup}

\noindent\textbf{Datasets and Protocols.}
We evaluate BiomedAP on 11 biomedical classification benchmarks covering diverse imaging modalities, including X-ray, MRI, dermoscopy, and fundus images \cite{pogorelov2017kvasir,yang2021medmnist}.
To comprehensively assess both adaptation quality and generalization, we follow two protocols.
Under the \textbf{few-shot} setting, each class is provided with $K\in\{1,2,4,8,16\}$ labeled samples, and we report the mean top-1 accuracy over 3 random seeds.
This protocol evaluates how effectively the model adapts under limited supervision.
Under the \textbf{base-to-novel} setting, classes are split into disjoint base and novel subsets, and performance is measured by base accuracy, novel accuracy, and their harmonic mean (HM).
This setting is designed to assess transferability to unseen classes after adaptation.

\noindent\textbf{Implementation.}
We use BiomedCLIP-PubMedBERT as the frozen backbone \cite{gu2021pubmedbert,zhang2023biomedclip}, and optimize only the learnable prompt parameters together with the lightweight fusion modules using AdamW on a single RTX 5090 (32GB).
Unless stated otherwise, gated fusion modules are inserted at layers $\{5,8\}$, which we found to offer a good trade-off between early semantic interaction and sufficient downstream refinement.
The anchor-related weights are selected to preserve stability on base classes while maintaining adaptation flexibility for novel classes.

\subsection{Main Results}

\noindent\textbf{Few-shot classification.}
Table~\ref{tab:avg_method} shows that BiomedAP consistently outperforms all baselines across every shot setting.
In the most challenging low-resource regime, BiomedAP reaches \textbf{63.57\%} at $K{=}1$, exceeding Biomed-DPT by \textbf{4.54} points.
The advantage remains stable as supervision increases, with BiomedAP achieving the best accuracy from $K{=}1$ through $K{=}16$ and reaching \textbf{75.09\%} at $K{=}16$.
These gains suggest that BiomedAP is not merely benefiting from stronger prompts, but from better grounded adaptation: intermediate cross-modal fusion strengthens lesion-relevant visual-semantic alignment, while the dual-anchor regularization stabilizes prompt learning when only a few labeled examples are available.

  \begin{table}[t]
  \centering
  \footnotesize 
  \setlength{\tabcolsep}{2.5pt} 
  \renewcommand{\arraystretch}{0.80} 
  \caption{\textbf{Few-shot results.} Avg accuracy (\%) over 11 datasets and 3 seeds.}
  \label{tab:avg_method}
  \begin{tabular}{lccccc}
  \toprule
  Method & $K=1$ & $K=2$ & $K=4$ & $K=8$ & $K=16$ \\
  \midrule
  \rowcolor{gray!30}\multicolumn{6}{c}{\textbf{Zero-shot}} \\
  BiomedCLIP~\cite{zhang2023biomedclip} & \multicolumn{5}{c}{42.05} \\
  BiomedCLIP+Ensemble & \multicolumn{5}{c}{56.02} \\
  \rowcolor{gray!30}\multicolumn{6}{c}{\textbf{Adapters}} \\
  CLIP-Adapter~\cite{gao2024clip} & 46.54 & 45.18 & 45.45 & 46.27 & 47.75 \\
  Tip-Adapter~\cite{zhang2022tip} & 49.11 & 54.06 & 58.93 & 63.46 & 67.18 \\
  \rowcolor{gray!30}\multicolumn{6}{c}{\textbf{Linear Probe}} \\
  Standard LP~\cite{radford2021learning} & 51.80 & 54.54 & 60.49 & 67.56 & 68.76 \\
  LP++~\cite{huang2024lp} & 53.57 & 53.55 & 57.26 & 64.89 & 68.70 \\
  \rowcolor{gray!30}\multicolumn{6}{c}{\textbf{Prompt Learning}} \\
  CoOp~\cite{zhou2022coop} & 50.18 & 54.17 & 59.77 & 65.85 & 69.72 \\
  CoCoOp~\cite{zhou2022conditional} & 48.53 & 51.28 & 54.69 & 61.09 & 65.10 \\
  KgCoOp~\cite{yao2023kgcoop} & 52.16 & 54.24 & 59.21 & 63.87 & 64.85 \\
  ProGrad~\cite{zhu2023prograd} & 51.37 & 54.40 & 60.61 & 65.50 & 67.03 \\
  BiomedCoOp~\cite{Koleilat_2025_CVPR} & 56.69 & 58.58 & 64.45 & 68.84 & 72.13 \\
  Biomed-DPT~\cite{peng2025biomeddpt} & 59.03 & 61.27 & 66.12 & 70.76 & 73.51 \\
  \rowcolor{blue!20}\textbf{BiomedAP (Ours)} & \textbf{63.57} & \textbf{64.89} & \textbf{67.92} & \textbf{71.21} & \textbf{75.09} \\
  \bottomrule
  \end{tabular}
  \end{table}

\begin{figure*}[t]
  \centering
  \includegraphics[width=1.0\textwidth]{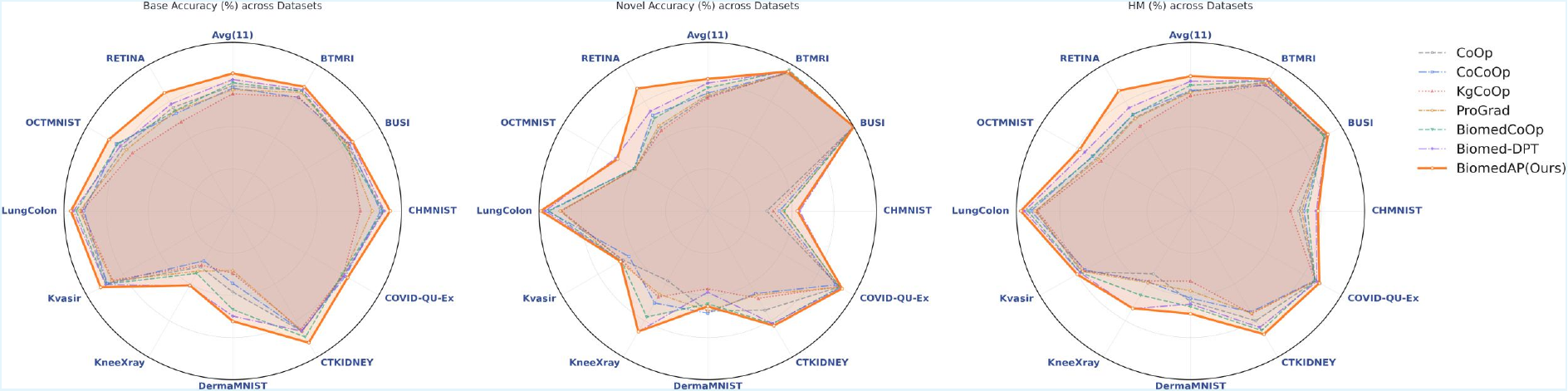}
  \caption{\textbf{Base-to-novel transfer.} Base/Novel accuracy and HM across 11 datasets.}
  \label{fig:radar}
\end{figure*}

\noindent\textbf{Base-to-novel transfer.}
BiomedAP further delivers the strongest base-to-novel generalization.
As summarized in Fig.~\ref{fig:radar}, it achieves the best average Base accuracy (\textbf{81.62\%}), Novel accuracy (\textbf{78.42\%}), and harmonic mean (\textbf{80.02\%}), improving over Biomed-DPT by \textbf{3.56} points on Base, \textbf{2.45} points on Novel, and \textbf{3.00} points on HM.
Notably, these gains arise from \emph{joint improvement} on both seen and unseen classes rather than from sacrificing one for the other.
This trend is especially clear on \textit{RETINA}, where BiomedAP raises Novel accuracy from 68.29\% to \textbf{83.94\%} and improves HM from 70.77\% to \textbf{82.49\%}.
We attribute this stronger transferability to confidence-aware optimization and dual-anchor regularization, which reduce overfitting to brittle expert-style prompts and yield more stable semantic adaptation under class shift.

\subsection{Robustness to Prompt Variations}
\label{sec:robustness}

\noindent\textbf{Setup.}
We perform \textit{inference-time perturbation analysis} \cite{zhou2022coop} by evaluating few-shot performance under five prompt templates (Table~\ref{tab:prompt_templates}), without retraining.

\begin{table}[t]
\centering
\scriptsize
\setlength{\tabcolsep}{3pt}
\renewcommand{\arraystretch}{0.90}
\caption{\textbf{Prompt templates for robustness evaluation.} ``\{c\}'' denotes the class name.}
\label{tab:prompt_templates}
\begin{tabular}{l p{0.33\columnwidth} p{0.27\columnwidth}}
\toprule
\textbf{Category} & \textbf{Template} & \textbf{Description} \\
\midrule
Minimal & ``\{c\}'' & Baseline format used in training \\
Article & ``a \{c\}'' & Minor syntactic variation \\
Generic & ``a photo of \{c\}'' & Out-of-domain natural-image style \\
Abstract & ``medical image of \{c\}'' & Domain-general but modality-agnostic \\
Extreme & ``'' (Empty) & Absence of textual context \\
\bottomrule
\end{tabular}
\end{table}

\begin{figure*}[t]
  \centering
  \begin{subfigure}[t]{0.50\textwidth}
    \centering
    \includegraphics[width=\textwidth]{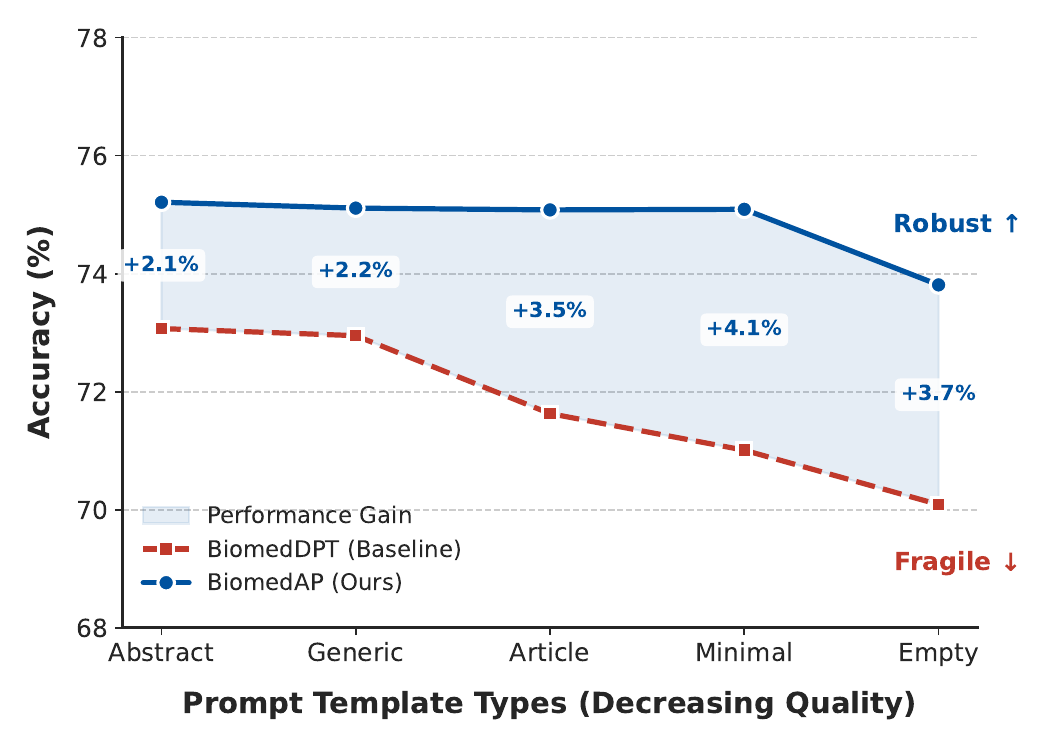}
    \caption{\textbf{Robustness to inference-time prompt variations.}}
    \label{fig:robustness_chart}
  \end{subfigure}\hfill
  \begin{subfigure}[t]{0.50\textwidth}
    \centering
    \includegraphics[width=\textwidth]{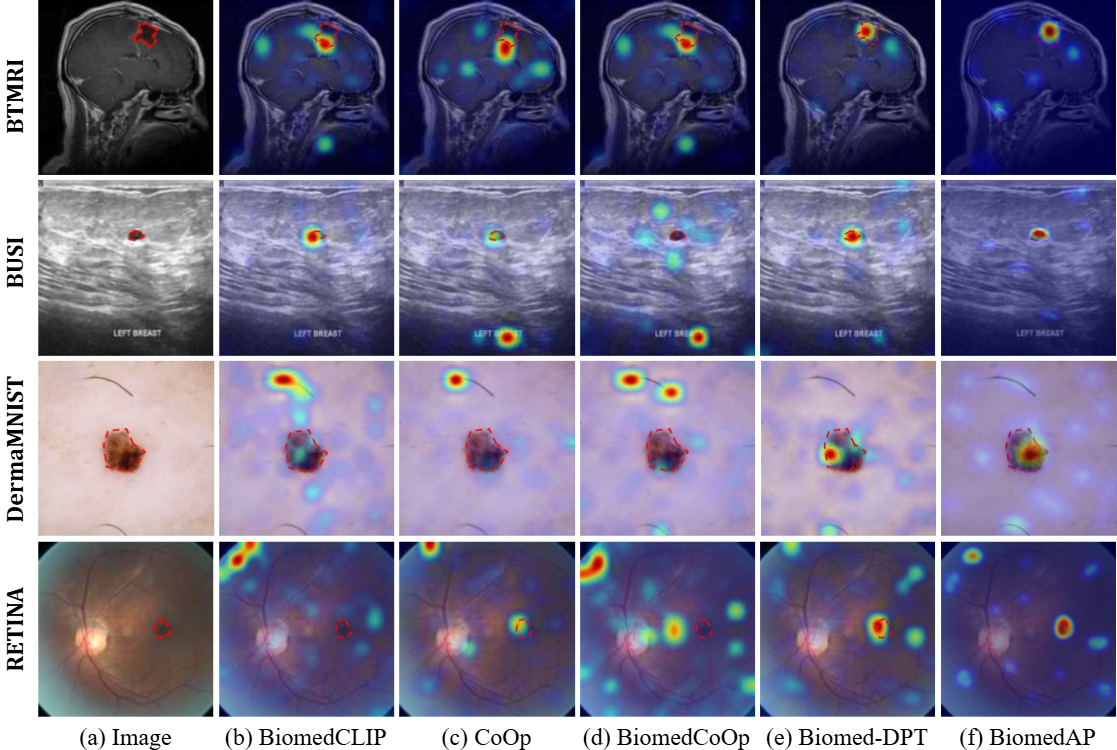}
    \caption{\textbf{Qualitative lesion localization.}}
    \label{fig:qual_cam}
  \end{subfigure}
  \caption{\textbf{Robustness and qualitative localization.}
  Left: few-shot performance under five prompt templates evaluated at inference time without retraining, ranging from the training-style minimal prompt to the extreme empty prompt.
  BiomedAP consistently surpasses BiomedDPT and shows smaller degradation under increasingly weak or noisy textual context.
  Right: representative activation maps for lesion localization.
  Compared with prior baselines, BiomedAP produces more compact and lesion-centered responses while suppressing diffuse activations over irrelevant regions.}
  \label{fig:robust_and_qual}
\end{figure*}

\noindent\textbf{Results.}
Fig.~\ref{fig:robustness_chart} shows that BiomedAP consistently outperforms BiomedDPT under all prompt variants, and that its advantage becomes more pronounced as prompt quality deteriorates.
Across the evaluated templates, BiomedAP yields a stable gain of 2.1--4.1 points, indicating substantially lower sensitivity to prompt formulation.
This trend is especially important in the low-quality regime: while BiomedDPT degrades more sharply under simplified or missing text, BiomedAP maintains stronger performance even in the extreme ``Empty'' setting, achieving 73.81\% versus 70.09\%.
These findings support our central claim that BiomedAP mitigates prompt fragility by reducing reliance on brittle prompt wording and preserving visually grounded semantics under inference-time text variation.

\subsection{Ablation Studies}

\noindent\textbf{Key components.}
Table~\ref{ablation} shows that the two core components of BiomedAP are individually effective and jointly complementary.
Starting from the baseline without GCPF or $\mathcal{L}_{anchor}$, adding GCPF raises the average few-shot accuracy from 62.14\% to 64.54\% and improves HM from 74.01\% to 74.94\%.
This result suggests that intermediate gated fusion alleviates modality isolation by enabling earlier and more grounded vision--text interaction during encoding.
Adding $\mathcal{L}_{anchor}$ yields further gains, increasing HM to 75.95\% and the few-shot average to 65.05\%, which indicates that anchor-based regularization stabilizes prompt semantics beyond pure fusion.
The full model achieves the strongest performance by a clear margin (HM: 80.02\%, Avg few-shot: 68.53\%), showing that cross-modal grounding and dual-anchor semantic stabilization address complementary failure modes in biomedical adaptation \cite{li2023blip2,li2022blip}.

\begin{table}[t]
\centering
\footnotesize
\caption{\textbf{Component ablation.} Impact of GCPF and $\mathcal{L}_{anchor}$.}
\label{ablation}
\setlength{\tabcolsep}{3.5pt}
\renewcommand{\arraystretch}{0.90}

\resizebox{\columnwidth}{!}{%
\begin{tabular}{lccccccccc}
\toprule
\multirow{2}{*}{Configuration} & \multicolumn{3}{c}{Base-to-Novel} & \multicolumn{6}{c}{Few-shot (Avg over $K$)} \\
\cmidrule(lr){2-4} \cmidrule(lr){5-10}
 & Base & Novel & HM & 1 & 2 & 4 & 8 & 16 & Avg \\
\midrule
w/o GCPF + w/o $\mathcal{L}_{anchor}$ & 77.21 & 70.81 & 74.01 & 55.15 & 57.16 & 61.87 & 66.80 & 69.72 & 62.14 \\
+ GCPF & 77.96 & 71.93 & 74.94 & 57.16 & 59.71 & 62.94 & 69.89 & 73.01 & 64.54 \\
+ $\mathcal{L}_{anchor}$ & 78.96 & 72.93 & 75.95 & 59.40 & 60.78 & 64.62 & 69.37 & 71.07 & 65.05 \\
\rowcolor{blue!10}\textbf{BiomedAP (Full)} & \textbf{81.62} & \textbf{78.42} & \textbf{80.02} & \textbf{63.57} & \textbf{64.89} & \textbf{67.92} & \textbf{71.21} & \textbf{75.09} & \textbf{68.53} \\
\bottomrule
\end{tabular}%
}
\end{table}

\begin{table}[!h]
\centering
\scriptsize
\setlength{\tabcolsep}{2.1pt} 
\renewcommand{\arraystretch}{0.84}

\caption{\textbf{Fusion layer sensitivity} (Acc., \%).}
\label{tab:fewshot_fusion_layers}

\begin{tabular}{@{}lccccc@{}}
\toprule
Method (Layers) & $K=1$ & $K=2$ & $K=4$ & $K=8$ & $K=16$ \\
\midrule
No fusion & 59.40 & 60.78 & 64.62 & 69.37 & 71.07 \\
Early-only ([2]) & 59.53 & 61.23 & 64.77 & 69.57 & 72.38 \\
Mid-only ([5]) & 60.91 & 62.61 & 65.54 & 70.17 & 73.37 \\
Late-only ([9]) & 60.53 & 61.42 & 64.64 & 69.11 & 73.15 \\
\rowcolor{blue!10}\textbf{Mid+Late ([5,8])} & \textbf{63.57} & \textbf{64.89} & \textbf{67.92} & \textbf{71.21} & \textbf{75.09} \\
Multi-layer Late ([8,9,10]) & 59.61 & 61.26 & 64.89 & 71.03 & 73.69 \\
\bottomrule
\end{tabular}

\end{table}

\begin{table}[t]
\centering
\small
\setlength{\tabcolsep}{6pt}
\renewcommand{\arraystretch}{0.92}
\caption{\textbf{Inference context comparison} (Accuracy, \%).}
\label{tab:inference_context}
\begin{tabular}{lcc}
\toprule
\textbf{Strategy} & \textbf{Mechanism} & \textbf{Accuracy} \\
\midrule
Null & No interaction & 71.67 \\
Retrieval (Top-1) & Hard selection & 73.78 \\
\rowcolor{blue!10}\textbf{Mean (Ours)} & \textbf{Global prior} & \textbf{75.09} \\
\midrule
\textit{Oracle (Ref.)} & \textit{Ground truth} & \textit{77.59} \\
\bottomrule
\end{tabular}
\end{table}

\noindent\textbf{Fusion placement.}
As shown in Table~\ref{tab:fewshot_fusion_layers}, placing fusion modules at both a mid and a late layer (\textbf{[5,8]}) consistently delivers the strongest results across all few-shot regimes, from $K{=}1$ to $K{=}16$.
This pattern indicates that effective cross-modal interaction benefits from a balance between \emph{semantic emergence} and \emph{refinement capacity}: fusion introduced too early operates on insufficiently abstract visual features, whereas fusion applied only at very late layers leaves little room for downstream correction.
Notably, the \textbf{Mid+Late} configuration also outperforms multi-layer late fusion ([8,9,10]), suggesting that diversity in fusion depth is more beneficial than simply increasing fusion frequency near the top of the encoder.

\noindent\textbf{Inference context.}
Table~\ref{tab:inference_context} further evaluates how fusion is conditioned at inference without access to ground-truth labels.
Among all label-free strategies, \textbf{Mean} performs best, reaching 75.09\% accuracy and narrowing the gap to the \textit{Oracle} reference (77.59\%).
It also surpasses \textbf{Retrieval (Top-1)} by 1.31 points and \textbf{Null} by 3.42 points, indicating that a global averaged prompt context provides a safer and more stable prior for class-agnostic fusion.
By contrast, hard retrieval may overcommit to a single candidate class and inject class-specific noise when the retrieved context is imperfect.

\noindent\textbf{Qualitative localization.}
Fig.~\ref{fig:qual_cam} provides qualitative evidence that BiomedAP improves cross-modal grounding across diverse biomedical modalities.
On representative examples from BTMRI, BUSI, DermaMNIST, and RETINA, BiomedAP produces more compact and lesion-centered activation maps, with the strongest responses concentrated around the annotated abnormal regions.
By contrast, prior baselines frequently show diffuse responses, multiple competing hotspots, or clear off-target activations outside the lesion area.
This trend is particularly notable in challenging cases where the lesion occupies only a small portion of the image, suggesting that BiomedAP better suppresses spurious textual guidance and preserves visually supported semantics.
These qualitative findings are consistent with our quantitative results, supporting that gated cross-modal fusion improves lesion-aware grounding and yields more robust predictions under heterogeneous clinical text.

\noindent\textbf{Impact of Vision-Language Backbones.}
Table~\ref{avg_backbone} evaluates BiomedAP on different pre-trained vision-language backbones \cite{radford2021learning,zhang2023biomedclip}.
General-domain CLIP performs worst, reaching only 24.44\% at $K{=}0$, which reflects the substantial mismatch between natural-image pre-training and biomedical classification \cite{gu2021pubmedbert}.
Although biomedical pre-training improves performance overall, \textbf{BiomedCLIP} consistently provides the strongest initialization and achieves the best accuracy across all shot settings from $K{=}0$ to $K{=}16$ \cite{eslami2021does,lin2023pmc,zhang2023biomedclip}.
These results indicate that BiomedAP benefits most from backbones with stronger domain-aligned vision-language priors, which can then be effectively adapted to downstream diagnostic tasks \cite{moor2023medflamingo,wang2022medclip}.

\begin{table}[t]
\centering
\scriptsize
\setlength{\tabcolsep}{3.8pt}
\renewcommand{\arraystretch}{0.92}
\caption{\textbf{Backbone comparison.} Classification accuracy (\%) using different pre-trained vision-language backbones. BiomedCLIP provides the strongest initialization for BiomedAP.}
\label{avg_backbone}
\resizebox{\columnwidth}{!}{%
\begin{tabular}{lcccccc}
\toprule
Backbone & $K{=}0$ & $K{=}1$ & $K{=}2$ & $K{=}4$ & $K{=}8$ & $K{=}16$ \\
\midrule
CLIP~\cite{radford2021learning} & 24.44 & 43.16 & 52.41 & 58.20 & 66.70 & 71.59 \\
PubMedCLIP~\cite{eslami2021does} & 27.28 & 48.07 & 55.93 & 59.97 & 65.48 & 68.47 \\
PMC-CLIP~\cite{lin2023pmc} & 20.92 & 36.35 & 43.09 & 49.37 & 54.14 & 56.80 \\
\textbf{BiomedCLIP}~\cite{zhang2023biomedclip} & \textbf{47.05} & \textbf{63.57} & \textbf{64.89} & \textbf{67.92} & \textbf{71.21} & \textbf{75.09} \\
\bottomrule
\end{tabular}%
}
\end{table}

\noindent\textbf{Sensitivity to Prompt Numbers.}
Table~\ref{dif_prompt} analyzes the effect of the number of LLM-generated prompts.
Expanding the prompt pool from 10 to 50 steadily improves performance across all shot settings; for instance, accuracy at $K{=}1$ increases from 56.45\% to 63.57\%.
This trend suggests that a more diverse set of textual descriptions provides broader clinical coverage and yields more informative supervision for prompt adaptation \cite{khattak2023maple,zhou2022coop}.
Because the gains become marginal beyond 50 prompts, we adopt 50 as the default setting to balance performance and efficiency.

\begin{table}[t]
\centering
\scriptsize
\setlength{\tabcolsep}{3.8pt}
\renewcommand{\arraystretch}{0.92}
\caption{\textbf{Effect of prompt quantity.} Ablation on the number of LLM-generated descriptions used for training. Performance improves with diversity and saturates around 50 prompts.}
\label{dif_prompt}
\resizebox{\columnwidth}{!}{%
\begin{tabular}{lcccccc}
\toprule
Prompts & $K{=}0$ & $K{=}1$ & $K{=}2$ & $K{=}4$ & $K{=}8$ & $K{=}16$ \\
\midrule
10 & 50.04 & 56.45 & 59.43 & 64.01 & 67.37 & 72.74 \\
20 & 53.24 & 57.32 & 60.59 & 64.50 & 68.97 & 73.42 \\
30 & 55.67 & 58.93 & 61.94 & 65.82 & 70.46 & 73.53 \\
40 & 55.08 & 61.98 & 62.44 & 66.27 & 70.58 & 73.70 \\
\textbf{50 (Default)} & \textbf{56.02} & \textbf{63.57} & \textbf{64.89} & \textbf{67.92} & \textbf{71.21} & \textbf{75.09} \\
\bottomrule
\end{tabular}%
}
\end{table}

\section{Conclusion}

We introduced BiomedAP, a parameter-efficient framework for robust biomedical vision-language adaptation under heterogeneous clinical text.
By addressing the limitations of modality isolation and prompt fragility, BiomedAP integrates gated intermediate-layer cross-modal fusion with dual-anchor semantic regularization, significantly improving robustness and adaptability for medical vision-language tasks.
Extensive experiments on 11 benchmarks show that BiomedAP outperforms existing methods in few-shot classification, base-to-novel generalization, and robustness against prompt perturbations, demonstrating its practical utility in clinical settings. Our findings highlight that robust adaptation relies on grounding visual evidence with stable semantic supervision, particularly in noisy or conflicting clinical text.
While qualitative evidence suggests improved lesion localization, future work will quantitatively assess this using metrics like the Dice coefficient.
BiomedAP adds modest computational overhead, making it suitable for medical PEFT, and future research will explore lighter fusion designs and extend the framework to 3D data and dense prediction tasks.

\section*{Acknowledgements}

This work was supported by the National Natural Science Foundation of China (U25A20450), the Zhejiang Provincial Natural Science Foundation (LQN26F020075), and the Humanities and Social Science Research Planning Fund of the Ministry of Education (20YJA790090).

The authors would like to thank kailiu for invaluable guidance and support throughout this project.

We also thank the anonymous reviewers for their constructive feedback, which helped improve the clarity and presentation of this work.

{
    \small
    \bibliographystyle{ieeenat_fullname}
    \bibliography{main}
}
\end{document}